# An Axiomatic Approach to Robustness in Search Problems with Multiple Scenarios

**Patrice Perny**
LIP6 - University of Paris VI
4 Place Jussieu
75252 Paris Cedex 05, France

**Olivier Spanjaard**
LAMSADE - University of Paris IX
Place du Maréchal de Lattre de Tassigny
75775 Paris Cedex 16, France

## Abstract

This paper is devoted to the the search of robust solutions in state space graphs when costs depend on scenarios. We first present axiomatic requirements for preference compatibility with the intuitive idea of robustness. This leads us to propose the Lorenz dominance rule as a basis for robustness analysis. Then, after presenting complexity results about the determination of robust solutions, we propose a new sophistication of $A^*$ specially designed to determine the set of robust paths in a state space graph. The behavior of the algorithm is illustrated on a small example. Finally, an axiomatic justification of the refinement of robustness by an OWA criterion is provided.

## 1 INTRODUCTION

In heuristic search in state space graphs, the value of an action allowing a transition between two nodes is usually represented by a scalar cost and the quality of a path by the sum of the costs of its arcs. In this framework, useful constructive search algorithms like $A^*$ and $A_\varepsilon^*$ (Hart et al., 1968; Pearl, 1984) have been proposed, performing the implicit enumeration of feasible solutions, directed by a numerical cost function to be minimized. However, in many practical search problems considered in Artificial Intelligence (e.g. path planning, game search, web search), the scalar-costs assumption in state space graphs does not fit, thus inducing additional sources of complexity for problem solving. For this reason, recent works use alternative assumptions and consider, for example, the problem of:

• *dealing with multiple criteria:* this concerns problems where the value of an action making a transition between two nodes must be evaluated according to different attributes, expressed on non-commensurate scales, non-necessarily reducible to a single additive cost function. In such problems, the state space graph representation with a vector-valued cost function is useful and leads to several possible extensions of $A^*$, e.g. MOA$^*$ to find all the efficient paths (Stewart and White III, 1991), U$^*$ to find a maximal path according to a multiattribute utility function, ABC$^*$ to find paths which best satisfies a set of prioritized soft constraints (Logan and Alechina, 1998).

• *dealing with uncertainty:* this concerns problems where the cost of the arcs are ill-known and characterized by uncertainty distributions. For example, when costs are time dependent and representable by random variables, the SDA$^*$ algorithm is used to determine the preferred paths according to the stochastic dominance partial order (Wellman et al., 1995). A sophistication of this algorithm specially designed to cope with both uncertainty and multiple criteria is proposed in Wurman and Wellman (1996).

In this paper we consider another variation of the search problem under uncertainty, that concerns situations where costs of paths might depend on different possible scenarios (states of the world) or come from discordant sources of information. In this context, our aim is to focus on the idea of *robustness* and the search of robust solutions as introduced in the following example:

**Example 1.** *We want to find the "best" path in the state space graph pictured on Figure 1 from a source node $a$ to a goal node $\gamma_1$ or $\gamma_2$ in a context where two different sets of costs must be considered.*

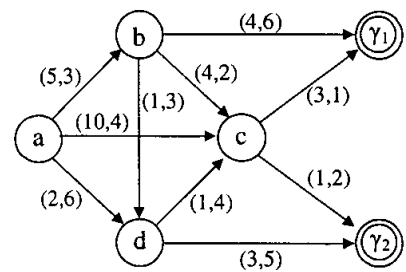

Figure 1: The state-space graph.

*This formal problem could be derived from different practical situations requiring decision making under uncertainty. For instance, consider an ambulance driver who wants to*

| Path | Nodes | Costs |
|---|---|---|
| 1 | $(a,b,\gamma_1)$ | (9,9) |
| 2 | $(a,b,c,\gamma_1)$ | (12,6) |
| 3 | $(a,b,c,\gamma_2)$ | (10,7) |
| 4 | $(a,b,d,c,\gamma_1)$ | (10,11) |
| 5 | $(a,b,d,c,\gamma_2)$ | (8,12) |
| 6 | $(a,b,d,\gamma_2)$ | (12,11) |
| 7 | $(a,c,\gamma_1)$ | (13,5) |
| 8 | $(a,c,\gamma_2)$ | (11,6) |
| 9 | $(a,d,c,\gamma_1)$ | (6,11) |
| 10 | $(a,d,c,\gamma_2)$ | (4,12) |
| 11 | $(a,d,\gamma_2)$ | (5,11) |

Table 1: The set of solution-paths.

*rush a man from point a to one of the city hospitals located in $\gamma_1$ and $\gamma_2$. Assume that two scenarios $s_1, s_2$ on the traffic in the city are considered, leading to a cost vector of type $(c(s_1), c(s_2))$ on each arrow. The problem is to determine the best destination and the best path. As an alternative example, consider a planning problem for an autonomous agent whose current objective is to reach one of the goal states $\gamma_1$ or $\gamma_2$ from the initial state a. The cost of each possible action is estimated by two external sensors located in different places and the agent does not know which is the most reliable sensor during the decision period. In both problems, one might be interested in finding a "robust" solution, i.e., a path which remains suitable whatever scenario (or sensor) is considered. This idea of robustness is consistent with the view of Kouvelis and Yu (1997) and Vincke (1999) but differs from the robustness considered in Aron and Van Hentenryck (2002) concerning minimal spanning tree problems with imprecise costs. The major difference is that, in our context, costs are linked to scenarios, thus making some combinations impossible. For example, considering Figure 1, the effective cost of path $(a,b,c)$ cannot be 7, because $(a,b)$ and $(b,c)$ cannot get simultaneously costs like 5 and 2 (or 3 and 4) respectively. Considering the graph pictured on Figure 1, the costs vectors of solution-paths are listed in Table 1.*

Facing such problems, simple scalarizations of cost-vectors do not lead to convincing results. For instance, using the average of the costs yields, among others, path 10 which is the worst solution if scenario $s_2$ occurs. Performing a weighted sum of the costs does not solve this problem either. Indeed, by geometrical arguments, it can easily be shown that solutions 1 and 3 cannot be obtained by minimizing a weighted sum of costs (they do not belong to the boundary of the convex hull of the points representing paths in the criteria space). Finally, focusing only on the worst cost over the scenarios (minimax criterion) is not really satisfactory due to overpessimistic evaluation. For example solution 3 cannot be obtained by the minimax criterion despite its promising costs due to the presence of the -indeed interesting- solution 1. Note that the dominance order is not more adequate since it yields too many solutions (paths 10, 11, 1, 3, 8, 7). These observations show the inadequacy of standard decision criteria to account for the idea of robustness as introduced above. Thus, the aim of the paper is:

• to propose an axiomatic framework for robustness and a formal definition of robust solutions,

• to introduce new algorithms to determine robust solutions in state space graphs.

The paper is organized as follows: Section 2 is devoted to the formal definition of robustness. Section 3 presents complexity results concerning the search of robust solutions and a heuristic search algorithm to find the set of robust paths in a state space graph. Finally in Section 4, we provide an axiomatic justification of the refinement of robustness by an OWA criterion.

## 2 DEFINING ROBUSTNESS

Considering a finite set of scenarios $S = \{s_1, \ldots, s_m\}$, any solution-path can be seen as an *act* $c : S \to \mathbb{R}_+$ in the sense of Savage (1954), characterized by the cost vector $(c(s_1), \ldots, c(s_m))$ in $\mathbb{R}_+^m$ whose $i^{th}$ component represents the cost of the path with respect to scenario $s_i$. Hence, the comparison of paths reduces to the comparison of their cost-vectors. In this framework, the following definitions are useful:

**Definition 1.** *The Weak-Pareto dominance relation (WP-dominance for short) on cost-vectors of $\mathbb{R}_+^m$ is defined, for all $x, y \in \mathbb{R}_+^m$ by:*

$$x \succsim_P y \iff [\forall i \in \{1, \ldots, m\}, x_i \leq y_i]$$

*The Pareto dominance relation (P-dominance for short) on cost-vectors of $\mathbb{R}_+^m$ is defined as the asymmetric part of $\succsim_P$:*

$$x \succ_P y \iff [x \succsim_P y \text{ and } not(y \succsim_P x)]$$

**Definition 2.** *Within a set $X$ any element $x$ is said to be P-dominated when $y \succ_P x$ for some $y$ in $X$, and P-non-dominated when there is no $y$ in $X$ such that $y \succ_P x$.*

In order to decide whether a path is better than another, we want to define a transitive preference relation $\succsim$ on cost-vectors capturing both the aim of cost-minimization and the idea of robustness. For this reason, the preference relation is expected to satisfy the following axioms:

**P-Monotonicity.** For all $x, y \in \mathbb{R}_+^m$, $x \succsim_P y \Rightarrow x \succsim y$ and $x \succ_P y \Rightarrow x \succ y$,

where $\succ$ is the strict preference relation defined as the asymmetric part of $\succsim$. This natural unanimity principle says that, if path $x$ has a lower cost than path $y$ whatever the scenario considered, then $x$ is preferred to $y$, and this preference is strict as soon as $x \neq y$. Then, the idea of robustness refers to equity in cost distribution among scenarios which can be expressed by the following axiom:

**Transfer Principle.** Let $x \in \mathbb{R}_+^m$ such that $x_i > x_j$ for some $i, j$. Then for all $\varepsilon$ such that $0 \leq \varepsilon \leq x_i - x_j$, $x - \varepsilon e_i + \varepsilon e_j \succsim x$ where $e_i$ (resp. $e_j$) is the vector whose $i^{th}$ (resp. $j^{th}$) component equals 1, all others being null.

This axiom captures the idea of robustness as follows: if $x_i > x_j$ for some cost-vector $x \in \mathbb{R}_+^m$, slightly improving (here decreasing) component $x_i$ to the detriment of $x_j$ while preserving the mean of the costs would produce a better distribution of costs, and consequently a more robust solution. Hence, path 1 should be at least as good as path 7 in Example 1 because there is an admissible transfer of size 4 between vectors (13, 5) and (9, 9). Note that using a similar transfer of size greater than 8 would increase inequality in terms of costs. This explains why the transfers must have a size $\varepsilon \leq x_i - x_j$. Such transfers are said to be *admissible* in the sequel. They are known as *Pigou-Dalton transfers* in Social Choice Theory, where they are used to reduce inequality in the income distribution over a population (see Sen (1997) for a survey).

Since elementary permutations of the vector $(x_1, \ldots, x_m)$ that just interchange two coordinates can be achieved using an admissible transfer, and since any permutation of $\{1, \ldots, n\}$ is the product of such elementary permutations, the Transfer Principle implies the following axiom:

**Symmetry.** For all $x \in \mathbb{R}_+^m$, for all permutations $\pi$ of $\{1, \ldots, m\}$, $(x_{\pi(1)}, \ldots, x_{\pi(m)}) \sim (x_1, \ldots, x_m)$,

where $\sim$ is the indifference relation defined as the symmetric part of $\succsim$. This axiom is natural in our context. Since no information about the likelihood of scenarios is available, they must be treated equivalently.

Note that the transfer principle possibly provides arguments to discriminate between vectors having the same average-cost but does not apply in the comparison of vectors having different average-costs. However, the possibility of discriminating is improved when combining the Transfer Principle with P-monotonocity. For example, consider paths 7 and 8 in Table 1 whose cost vectors are $(13, 5)$ and $(11, 6)$ respectively. Although P-dominance cannot discriminate between these two vectors, the discrimination is possible for any preference relation $\succsim$ satisfying both the Transfer Principle and the P-monotonocity axiom. Indeed, on the one hand, $(11, 6) \succ_P (12, 6)$ and therefore $(11, 6) \succ (12, 6)$ thanks to P-monotonicity; on the other hand, $(12, 6) \succsim (13, 5)$ thanks to the Transfer Principle applied to the transfer $(13 - 1, 5 + 1) = (12, 6)$. Hence, we get: $(11, 6) \succ (13, 5)$ by transitivity. In order to better characterize those vectors that can be compared using such combination of the P-monotonicity and the Transfer Principle we recall the definition of Lorenz vectors and related concepts (for more details see e.g. Marshall and Olkin (1979); Shorrocks (1983)):

**Definition 3.** *For all* $x \in \mathbb{R}_+^m$, *the* Generalized Lorenz Vector *associated to $x$ is the vector:*

$$L(x) = (x_{(1)}, x_{(1)} + x_{(2)}, \ldots, x_{(1)} + x_{(2)} + \ldots + x_{(m)})$$

where $x_{(1)} \geq x_{(2)} \geq \ldots \geq x_{(m)}$ represents the components of $x$ sorted by decreasing order. The $k^{th}$ component of $L(x)$ is $L_k(x) = \sum_{i=1}^{k} x_{(i)}$.

**Definition 4.** *The* Generalized Lorenz dominance relation *(L-dominance for short) on $\mathbb{R}_+^m$ is defined by:*

$$\forall x, y \in \mathbb{R}_+^m, \ x \succsim_L y \iff L(x) \succsim_P L(y)$$

The notion of Lorenz dominance was initially introduced to compare vectors with the same average cost and its link to the transfer principle was established by Hardy et al. (1934). The generalized version of L-dominance considered here is classical (see e.g. Marshall and Olkin (1979)) and allows any pair of vectors in $\mathbb{R}_+^m$ to be compared. Within a set $X$, any element $x$ is said to be *L-dominated* when $y \succ_L x$ for some $y$ in $X$, and *L-non-dominated* when there is no $y$ in $X$ such that $y \succ_L x$. In order to establish the link between Generalized Lorenz dominance and preferences satisfying combination of P-Monotonocity and Transfer Principle we recall a result of Chong (1976):

**Theorem 1.** *For any pair of distinct vectors $x, y \in \mathbb{R}_+^m$, if $x \succsim_P y$, or if $x$ obtains from $y$ by a Pigou-Dalton transfer, then $x \succsim_L y$. Conversely, if $x \succsim_L y$, then there exists a sequence of admissible transfers and/or Pareto-improvements to transform $y$ into $x$.*

This theorem establishes $\succsim_L$ as the minimal transitive relation (with respect to set inclusion) satisfying simultaneously P-Monotonicity and the Transfer Principle. As a consequence, the subset of L-non-dominated elements appears as a very natural solution to choice problems with multiple scenarios, as far as robustness is concerned. For this reason, we investigate in the next section the generation of the set of L-non-dominated paths in a state space graph.

## 3 SEARCH FOR ROBUST SOLUTIONS

### 3.1 COMPUTATIONAL COMPLEXITY

We investigate here the computational complexity of the search of the set of L-non-dominated solution-paths in a state space graph. Note first that the L-non-dominated solutions are a subset of the P-non-dominated solutions which might be very numerous. We wish to evaluate the extend to which focusing on L-non-dominated solutions (rather than P-non-dominated solutions) reduces the size of the solution space. In this respect, the study of the pathological instance introduced in Hansen (1980) for the multi-objective shortest path problem is quite significant (one looks for the set of P-non-dominated paths from a source node to a destination node, see Figure 2). In that bivalued graph, all the paths from node 1 to node $2p + 1$ have the same average-cost (whose value is $(2^p - 1)/2$) but distinct costs on the first component (due to the uniqueness of the binary representation of an integer). The resulting set of cost-vectors is $\{(x, 2^p - 1 - x), x \in \{0, \ldots, 2^p - 1\}\}$, which contains only

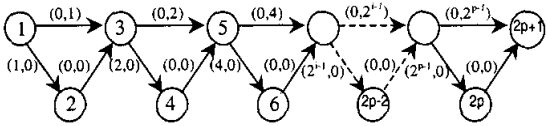

Figure 2: The pathological instance of Hansen.

P-non-dominated elements by construction. Notice that the cardinal of this set is exponential in the size of the graph. However, due to the Transfer Principle, there exists only two L-non-dominated cost vectors (those minimizing the difference between their components). Unfortunately, it is also possible to exhibit pathological instance for our problem, such as the graph on Figure 3 (where every arc without cost-vectors is actually valued $(0,0)$). Indeed, all the paths from node 0 to node $2p+1$ have distinct Lorenz vectors and are L-non-dominated. The proof is similar to the previous one. The set of cost vectors associated with the solution-paths of the graph is $\{(2x, 3 \times 2^p - x), x \in \{0, \ldots, 2^p - 1\}\}$. Note that the second component is always greater than the first component for $x \in \{0, \ldots, 2^p - 1\}$. Consequently, the corresponding set of Lorenz vectors writes $\{(3 \times 2^p - x, 3 \times 2^p + x), x \in \{0, \ldots, 2^p - 1\}\}$. All Lorenz vectors have the same average-cost and distinct values on the the first component. Moreover, the size of that set is exponential in the size of the graph. Due to the potentially

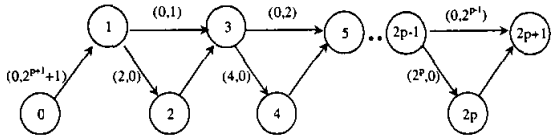

Figure 3: An instance where all paths are L-non-dominated.

exponential number of L-non-dominated paths, the following proposition is immediate:

**Proposition 1.** *The problem of finding L-non-dominated paths in a graph is, in worst case, intractable, i.e. requires for some problems a number of operations which grows exponentially with the size of the problem.*

In other respects, one may be interested in the complexity of deciding whether there exists a path whose cost distribution L-dominates a given cost-vector. The following result establishes that this decision problem cannot be solved in polynomial time unless $P = NP$:

**Proposition 2.** *Deciding whether there exists a path whose cost distribution L-dominates a given cost-vector is an NP-complete decision problem.*

**Proof.** We reduce the partition problem to our problem.

*instance:* Finite set $A = \{a_1, \ldots, a_p\}$ and a size $s(a) \in \mathbb{Z}^+$ for each $a \in A$.

*question:* Is there a subset $A' \subseteq A$ such that $\sum_{a \in A'} s(a) = \sum_{a \in A - A'} s(a)$?

That problem is proved NP-complete (see e.g. Garey and Johnson (1979)). One constructs -in polynomial time- a graph as indicated on Figure 4 (where every arc without cost-vectors is actually valued $(0,0)$). Deciding whether there exists a path from node 1 to node $2p+1$ such that its vector-cost L-dominates the vector $(\frac{\sum_{a \in A} s(a)}{2}, \frac{\sum_{a \in A} s(a)}{2})$ amounts to solve the partition problem. □

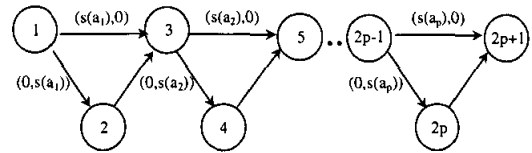

Figure 4: Reduction from partition problem.

### 3.2 ALGORITHM

Due to the existence of multiple scenarios, the search of a robust solution can be seen as a particular specification of a multi-objective path problems (the objectives corresponding to the different scenarios). Several variations of A* have been studied to generate the P-non-dominated solution-paths in multi-objective problems, see e.g. Stewart and White III (1991); Dasgupta et al. (1996a). As the set of L-non-dominated solution-paths is included in the set of P-non-dominated solution-paths, we can use a sophistication of these algorithms that exploits the exact nature of L-dominance (using the idea of approximation of a preference relation suggested in Perny and Spanjaard (2002)).

Let us briefly recall some essential features of multi-objective A*. In a graph valued by cost-vectors, there possibly exists several P-non-dominated paths to reach a given node. Hence, at each node $n$, one stores a *set $G(n)$* of cost-vectors $g(n)$ corresponding to P-non-dominated paths arriving in $n$. Moreover, a node $n$ may be on the path of more than one P-non-dominated solutions. Consequently, a *set $H(n)$* of heuristic cost-vectors $h(n)$ is assigned to each node $n$. Finally, at each node $n$, a *set $F(n)$* of evaluation vectors $f(n)$ is computed from all possible combinations $\{g(n) + h(n) : g(n) \in G(n), h(n) \in H(n)\}$.

The algorithm we propose to compute robust solution-paths relies on this general pattern. Before presenting our search algorithm itself, we establish two preliminary results on which our pruning rule and our priority rule are grounded:

**Proposition 3.** *For any $x, y, z \in \mathbb{R}_+^m$, $[x \succ_L y$ and $y \succ_P z] \Longrightarrow x \succ_L z$*

**Proof.** By definition $y \succ_P z \Longrightarrow y \succ_L z$. Since $x \succ_L y$ we get $x \succ_L z$ by transitivity. □

**Proposition 4.** *A cost-vector* $(x_1, \ldots, x_m)$ *L-dominates any cost-vector* $(y_1, \ldots, y_m)$ *such that* $\sum_{i=1}^{m} y_i > m.x_{(1)}$.

**Proof.** Assume that: $(i)$ $\sum_{i=1}^{m} y_{(i)} > m.x_{(1)}$ and $(ii)$ $\exists k \leq m$ s.t. $\sum_{i=1}^{k} y_{(i)} < \sum_{i=1}^{k} x_{(i)}$. We know that $y_{(1)} > x_{(1)}$ thanks to $(i)$. Due to $(ii)$, $\exists j \in \{1, \ldots, k\}$ s.t. $y_{(j)} < x_{(j)} \leq x_{(1)}$ and therefore $y_{(k)} < x_{(1)}$ (because $y_{(k)} \leq y_{(j)}$). Then, we have $\sum_{i=1}^{k} x_{(i)} \leq k.x_{(1)}$ which implies $(iii)$ $\sum_{i=1}^{k} y_{(i)} < k.x_{(1)}$ by $(ii)$. Moreover, we have $(iv)$ $\sum_{i=k+1}^{m} y_{(i)} \leq (m-k)y_{(k)}$. Since $y_{(k)} \leq y_{(j)} < x_{(1)}$ we have $(m-k)y_{(k)} < (m-k)x_{(1)}$ and therefore we get $(v)$ $\sum_{i=k+1}^{m} y_{(i)} < (m-k)x_{(1)}$ by $(iv)$. Finally we get: $\sum_{i=1}^{m} y_i < m.x_{(1)}$ by $(iii)$ and $(v)$ which yields a contradiction. □

Here are the main features of our algorithm, where we expand labels rather than nodes:

**Output:** we determine the set of L-non-dominated paths from the source node to a goal node. If several paths have the same L-non-dominated Lorenz vector, we detect only one path among them.

**Heuristics:** like in MOA*, we use an admissible set $H(n)$ of vector-valued heuristic costs, i.e. for any cost-vector $c$ of a P-non-dominated path from $n$ to a goal node, there exists $h(n) \in H(n)$ such that $h(n) \succsim_P c$.

**Priority:** the search is totally ordered by a lexicographic order on evaluation vectors $f(n)$ defined at each open node $n$. This evaluation function $f(n)$ is obtained in two steps. We first compute $F(n)$, the set of all costs vectors of type $g(n) + h(n)$ where $g(n)$ is the cost-vector of any P-non-dominated subpath arriving in $n$ and $h(n)$ is the cost of any vector in the heuristic set $H(n)$. Then, $f(n)$ is defined as the best element of the set $\{L(x), x \in F(n)\}$ using the following lexicographic order:

$$L(x) \succ_{\text{lex}} L(y) \iff \exists k \begin{cases} \forall i < k : L_i(x) = L_i(y) \\ L_k(x) < L_k(y) \end{cases}$$

This is also the lexicographic order used to rank the open nodes by decreasing order of priority according to their evaluation vector $f(n)$. The choice of that priority rule is motivated by two remarks:

- a heuristic consideration: the early detection of a minimax-optimal solution-path[1] potentially speeds up the search by providing the best bound to prune paths according to Proposition 4.

- a prudence consideration: at the goal nodes, such a priority rule guarantees to expand only labels corresponding to L-non-dominated solution-paths.

**Stopping condition:** the algorithm is kept running until there is no remaining subpath able to reach a new L-non-dominated solution-path, i.e. all open labels are either P-dominated by another label on the same node or L-dominated by a solution-path already detected.

**Pruning:** the Bellman principle does not hold for L-dominance i.e. a L-non-dominated path could contain a L-dominated subpath. For example, assume that two subpaths $P_1$ and $P_2$ of costs $(3, 2)$ and $(1, 4)$ lead to the same node. It is easy to see that $P_1 \succ_L P_2$. However, if we extend both subpaths by the same subpath $P_3$ of cost $(3, 1)$, then $P_2 \cup P_3 \succ_L P_1 \cup P_3$. That is why, we must be very careful in pruning L-dominated subpaths during the search. Consequently, we use the two following pruning rules, that we apply at the beginning of each iteration:

- we prune any subpath whose value of the evaluation function is L-dominated by (or equal to) an already detected solution-path;
- we prune any subpath P-dominated by (or equal to) another subpath at the same node, as it is usual in multiobjective heuristic search.

Such subpaths cannot lead to new L-non-dominated cost vectors at a goal node. Indeed, as soon as we assume there exists a strictly positive lower bound on each cost, a path is P-dominated by any of its subpaths. Consequently, by Proposition 3, if a subpath is L-dominated by an already detected solution-path, any extension will be also L-dominated by the same solution-path. Similarly, if a subpath is P-dominated by another subpath at the same node, any extension will be also P-dominated (thus L-dominated) by the corresponding extension of the P-dominating subpath. For the convenience of the reader, we give now a detailed example to illustrate the behavior of the algorithm.

**Example 2.** *Consider the graph of Figure 1. The arc costs are shown beside each arc. The heuristic set at a node $n$ is defined by the P-non-dominated cost vectors of the arcs with tails at node $n$. Obviously, such vectors under-estimate the remaining costs to reach a goal node, thus leading to an admissible heuristic. For instance, the set of heuristics at node $a$ is $\{(5, 3), (2, 6)\}$. Let $[n, g, L(f)]_p$ denote a label of a path from the source to node $n$, where $g$ is the cost-vector of that path, $L(f)$ is the Lorenz vector corresponding to the value $f$ of the evaluation function and $p$ is a pointer to the previous node along the path. The trace of the algorithm is indicated on Table 2, with the following conventions: * pinpoints the pruned labels whereas ☆ pinpoints the L-non-dominated solution-paths. The pruning rules speed up the exploration of the state space graph, as shown hereafter. At iteration 3, label $[d, (6, 6), (10, 17)]_d$ is pruned since $(6, 6)$ is P-dominated by $(2, 6)$ (three solutions-paths avoided). At iteration 6, label $[\gamma_1, (12, 6), (12, 18)]_c$ is pruned since its evaluation is L-dominated by $(9, 18)$. At iteration 7, labels $[c, (10, 4), (11, 17)]_a$ and $[c, (3, 10), (11, 17)]_d$ are pruned since their evaluations are L-dominated by $(10, 17)$ (four solution-paths avoided). At iteration 8, there is no*

---

[1] Note that the greatest component of a cost vector is the first component of the corresponding Lorenz vector.

| iteration | open labels | expanded label |
|---|---|---|
| 1 | $[a,(0,0),(5,8)]$ | $[a,(0,0),(5,8)]$ |
| 2 | $[b,(5,3),(6,12)]_a$<br>$[c,(10,4),(11,17)]_a$<br>$[d,(2,6),(10,13)]_a$ | $[b,(5,3),(6,12)]_a$ |
| 3 | $[c,(10,4),(11,17)]_a$<br>$[d,(2,6),(10,13)]_a$<br>$[\gamma_1,(9,9),(9,18)]_b$<br>$[c,(9,5),(10,17)]_b$<br>$[d,(6,6),(10,17)]_b*$ | † $[\gamma_1,(9,9),(9,18)]_b$ |
| 4 | $[c,(10,4),(11,17)]_a$<br>$[d,(2,6),(10,13)]_a$<br>$[c,(9,5),(10,17)]_b$ | $[d,(2,6),(10,13)]_a$ |
| 5 | $[c,(10,4),(11,17)]_a$<br>$[c,(9,5),(10,17)]_b$<br>$[c,(3,10),(11,17)]_d$<br>$[\gamma_2,(5,11),(11,16)]_d$ | $[c,(9,5),(10,17)]_b$ |
| 6 | $[c,(10,4),(11,17)]_a$,<br>$[c,(3,10),(11,17)]_d$<br>$[\gamma_2,(5,11),(11,16)]_d$<br>$[\gamma_1,(12,6),(12,18)]_c*$<br>$[\gamma_2,(10,7),(10,17)]_c$ | † $[\gamma_2,(10,7),(10,17)]_c$ |
| 7 | $[c,(10,4),(11,17)]_a*$<br>$[c,(3,10),(11,17)]_d*$<br>$[\gamma_2,(5,11),(11,16)]_d$ | † $[\gamma_2,(5,11),(11,16)]_d$ |
| 8 | ∅ | stop |

Table 2: Trace of the algorithm.

*more label and the algorithm stops. In all, the pruning rules enable here to break the exploration of seven L-dominated solution-paths (over eight) before they reach a goal node. In other respects, note that the expansion of labels rather than nodes allows the expansion of labels $[c,(10,4),(11,17)]_a$ and $[c,(3,10),(11,17)]_d$ to be avoided.*

## 4 REFINING ROBUSTNESS

As shown in Subsection 3.1, the set of L-non-dominated solutions is a subset of P-non-dominated solutions but it might contain an important number of elements. For this reason, we would like to refine the notion of robust solution by proposing a criterion allowing to discriminate between the L-non-dominated solutions. We propose here an axiomatic result concerning the numerical representation of a preference weak-order $\succsim$ on $X = \mathbb{R}_+^m$ consistent with L-dominance.

The first axiom requires that the only relevant information to discriminate between solutions is the corresponding generalized Lorenz vector:

**Neutrality.** For all $x, y$ in $X$, $L(x) = L(y) \Rightarrow x \sim y$.

We may define a preference relation $\succsim'$ among Lorenz vectors of $L(X) = \{v \in \mathbb{R}_+^m : \exists x \in \mathbb{R}_+^m, v = L(x)\}$ by setting, for any $L, M, \in L(X)$

$$L \succsim' M \Leftrightarrow \exists x, y \in X, \begin{cases} L(x) = L \text{ and } L(y) = M \\ x \succsim y \end{cases}$$

For the sake of convenience, we now use $\succsim$ instead of $\succsim'$ to denote the preference relation among Lorenz vectors. As we intend the preference relation to refine L-dominance, we need the following axiom:

**Strict L-Monotonicity.** $L(x) \succ_L L(y) \Rightarrow x \succ y$.

Then we introduce three axioms that can be seen as counterparts of von Neumann and Morgenstern (1947) axioms adapted for Lorenz vectors.

**Complete weak-order.** $\succsim$ is reflexive, transitive and complete.

**Continuity.** Let $L, M, N \in L(X)$ such that $L \succ M \succ N$. There exists $\alpha, \beta \in ]0, 1[$ such that:

$$\alpha L + (1 - \alpha)N \succ M \succ \beta L + (1 - \beta)N$$

**Independence.** Let $L, M, N$ belong to $L(X)$. Then, for all $\alpha \in ]0, 1[$:

$$L \succ M \Longrightarrow \alpha L + (1 - \alpha)N \succ \alpha M + (1 - \alpha)N$$

It is important to observe that this independence axiom is a weakening of the usual independence axiom on $X$, obtained by restriction to comonotonic vectors, where $x$ and $y$ in $X$ are said to be *comonotonic* if $x_i > x_j$ and $y_i < y_j$ for no $i, j \in \{1, \ldots, m\}$. Indeed, for any pair $x, y$ of comonotonic vectors, there exists a permutation $\pi$ of $\{1, \ldots, m\}$ such that $x_{\pi(1)} \geq x_{\pi(2)} \geq \ldots \geq x_{\pi(m)}$ and $y_{\pi(1)} \geq y_{\pi(2)} \geq \ldots \geq y_{\pi(m)}$. Consequently, $L(\alpha x + (1 - \alpha)y) = \alpha L(x) + (1 - \alpha)L(y)$. Hence, for all comonotonic vectors $x, y, z \in X$, if $x \succ y \Longrightarrow \alpha x + (1 - \alpha)z \succ \alpha y + (1 - \alpha)z$ then $L(x) \succ L(y) \Longrightarrow \alpha L(x) + (1 - \alpha)L(z) \succ \alpha L(y) + (1 - \alpha)L(z)$. Observing that for any triple $L, M, N$ of Lorenz vectors, there exists $x, y, z$, three comonotonic vectors in $X$ such that $L = L(x), M = L(y)$ and $N = L(z)$, we deduce that usual independence on $X$ implies independence on $L(X)$.

Note that weakening the usual independence is necessary in our framework due to its incompatibility with the Strict L-monotonicity axiom, as shown by the following:

**Example 3.** *Let us consider $x = (24, 24)$, $y = (22, 26)$ and $z = (26, 22)$. Due to Strict L-monotonicity $x \succ y$. Hence, usual independence would imply $(25, 23) = \frac{1}{2}x + \frac{1}{2}z \succ \frac{1}{2}y + \frac{1}{2}z = (24, 24)$ which is in contradiction with $(24, 24) \succ_L (25, 23)$.*

The conflict here can be explained as follows: on the one hand, the cost-dispersion of vector $(25, 23)$ resulting from the combination of $x$ and $z$ is greater than that of $x = (24, 24)$; on the other hand, the cost dispersion of vector $(24, 24)$ resulting from the combination of $y$ and $z$ is smaller than that of $y = (22, 26)$. This situation cannot occur when $x$, $y$ and $z$ are pairwise comonotonic, which explains the very idea of our independence axiom.

Actually, a similar idea was already present in Dual Choice Theory under Risk (see Yaari (1987)) in the form of the *Dual independence axiom*. The link with Yaari's theory under Risk is natural here since Lorenz vectors can be seen as counterparts of cumulative distribution functions in decision under risk.

Before introducing our representation theorem, we need to show that $L(X)$ with the usual convex combination in vector spaces is a *mixture set* (Herstein and Milnor, 1953):

**Definition 5.** *A mixture set is a set $\mathcal{M}$ and a function $f$ that assigns an element $f(\alpha, x, y) = \alpha x + (1 - \alpha)y$ in $\mathcal{M}$ to each $\alpha$ in $[0, 1]$ and each ordered pair $(x, y)$ in $\mathcal{M} \times \mathcal{M}$ such that:*

*M1.* $1x + 0y = x$,
*M2.* $\alpha x + (1 - \alpha)y = (1 - \alpha y) + \alpha x$,
*M3.* $\alpha[\beta x + (1 - \beta)y] + (1 - \alpha)y = (\alpha\beta)x + (1 - \alpha\beta)y$,

*for all $x, y$ in $\mathcal{M}$ and $\alpha, \beta$ in $[0, 1]$.*

We have:

**Lemma 1.** *$L(X)$ is a mixture set with respect to the usual convex combination in vector spaces.*

**Proof.** We first establish that $\alpha L + (1 - \alpha)M$ belongs to $L(X)$. Consider two vectors $x$ and $y$ in $X$ such that $L = L(x)$ and $M = L(y)$. It is easy to check that $\alpha L + (1 - \alpha)M = L(\alpha x + (1 - \alpha)y)$ and therefore $\alpha L + (1 - \alpha)M \in L(X)$. Then, M1 and M2 being straightforward, we only prove M3: $\alpha[\beta L + (1 - \beta)M] + (1 - \alpha)M = \alpha\beta L + \alpha M - \alpha\beta M + M - \alpha M = \alpha\beta L + (1 - \alpha\beta)M$. $\square$

A linear function on a mixture set is defined as follows:

**Definition 6.** $\varphi : \mathcal{M} \to \mathbb{R}$ *is linear if $\varphi(\alpha x + (1 - \alpha)y) = \alpha\varphi(x) + (1 - \alpha)\varphi(y)$ for all $\alpha \in [0, 1]$ and $x, y \in M$.*

Note that here, since the mixture operation coincides with the usual convex combination in vector spaces, $\varphi$ is automatically $m$-linear: $\varphi(\sum_{i=1}^{m} \alpha_i x_i) = \sum_{i=1}^{m} \alpha_i \varphi(x_i)$ with $\alpha_i \in [0, 1]$ for all $i$ (proof by recursion).

Moreover, note that the set $\{\ell_i = (1, 2, \ldots, i-1, i, \ldots, i) : i = 1, \ldots, m\}$ is a generator set for $L(X)$ (every element can be seen as a combination of those vectors of $L(X)$). Indeed, by setting $\ell_0 = (0, \ldots, 0)$ and $\ell_{m+1} = \ell_m$, we can write $e_i = 2\ell_i - \ell_{i-1} - \ell_{i+1}$ for all $i$ in $\{1, \ldots, m\}$, where $e_i$ is the vector whose $i^{th}$ component equals 1, all others being null. Consequently, every vector $L$ of $L(X)$ can be written:

$$L = \sum_{i=1}^{m} L_i e_i = \sum_{i=1}^{m} L_i(2\ell_i - \ell_{i-1} - \ell_{i+1}) = \sum_{i=1}^{m}(2L_i - L_{i-1} - L_{i+1})\ell_i \quad (1)$$

with the convention $L_0 = 0$ and $L_{m+1} = L_m$. We can now establish our representation theorem:

**Theorem 2.** *A preference relation $\succsim$ satisfies Neutrality, Strict L-monotonicity, Complete weak-order, Continuity and Independence iff there is a linear function $\varphi$ on $L(X)$ such that $x \succsim y \iff \varphi(L(x)) \leq \varphi(L(y))$ where:*

$$\varphi(L(x)) = \sum_{i=1}^{m}(2\varphi(\ell_i) - \varphi(\ell_{i-1}) - \varphi(\ell_{i+1}))L_i(x)$$

*and $\varphi(\ell_i) - \varphi(\ell_{i-1}) > \varphi(\ell_{i+1}) - \varphi(\ell_i) > 0$ for all $i$*

**Proof.** By Neutrality, $x \succsim y$ iff $L(x) \succsim L(y)$ and therefore assuming a complete weak-order on $X$ amounts to assuming a complete weak-order on $L(X)$. Using the classical result of Herstein and Milnor (1953) on mixture sets and Lemma 1, the following two statements are equivalent:

- Complete weak-order, Continuity and Independence hold;
- there is a linear function $\varphi$ on $L(X)$ that preserves $\succ$: for all $L, M \in L(X)$, $L \succ M$ iff $\varphi(L) < \varphi(M)$.

For every vector $L(x)$ of $L(X) \setminus \{\ell_0\}$ we have $2L_i(x) - L_{i-1}(x) - L_{i+1}(x) = x_{(i)} - x_{(i+1)} \geq 0$ for $i = 1, \ldots, m$ with the convention $x_{m+1} = 0$. Hence $\sum_{i=1}^{m}(L_i(x) - L_{i-1}(x) - L_{i+1}(x)) = \sum_{i=1}^{m} x_{(i)} - \sum_{i=1}^{m} x_{(i+1)} = x_{(1)}$. Then the coefficients $(L_i(x) - L_{i-1}(x) - L_{i+1}(x))/x_{(1)}$ are positive and add-up to 1. By the $m$-linearity of $\varphi$, $\varphi(\ell_0) = 0$ and for every vector $L(x)$ of $L(X) \setminus \{\ell_0\}$ we deduce thanks to Equation 1: $\varphi(L(x)/x_{(1)}) = \varphi(\sum_{i=1}^{m}[(2L_i(x) - L_{i-1}(x) - L_{i+1}(x))/x_{(1)}]\ell_i) = \sum_{i=1}^{m}[(2L_i(x) - L_{i-1}(x) - L_{i+1}(x))/x_{(1)}]\varphi(\ell_i) = 1/x_{(1)} \sum_{i=1}^{m}(2L_i(x) - L_{i-1}(x) - L_{i+1}(x))\varphi(\ell_i)$. Then multiplication by $x_{(1)}$ yields $\varphi(L(x)) = \sum_{i=1}^{m}(2L_i(x) - L_{i-1}(x) - L_{i+1}(x))\varphi(\ell_i) = \sum_{i=1}^{m}(2\varphi(\ell_i) - \varphi(\ell_{i-1}) - \varphi(\ell_{i+1}))L_i(x)$.

Moreover, Strict L-monotonicity implies that $2\varphi(\ell_i) > \varphi(\ell_{i+1}) + \varphi(\ell_{i-1})$ since $\ell_{i+1} + \ell_{i-1} \succ_L 2\ell_i$, and $\varphi(\ell_{i+1}) > \varphi(\ell_i)$ since $\ell_i \succ_L \ell_{i+1}$. Conversely, if $\varphi(\ell_i) - \varphi(\ell_{i-1}) > \varphi(\ell_{i+1}) - \varphi(\ell_i) > 0$ for all $i \in \{1, \ldots, m\}$, then Strict L-Monotonicity clearly holds. This concludes the proof. $\square$

The linear function $\varphi$ on $L(X)$ can also be written directly on $X$ as follows:

$$\varphi(x) = \sum_{i=1}^{m}(\varphi(\ell_i) - \varphi(\ell_{i-1}))x_{(i)}$$

for all $x$ in $X$. We recognize an Ordered Weighted Average (OWA, Yager, 1998) with strictly decreasing and strictly positive weights $w_i = \varphi(\ell_i) - \varphi(\ell_{i-1})$. This is consistent with a result in Ogryczak (2000), where it is shown that any solution minimizing an ordered weighted average with strictly decreasing and strictly positive weights is L-nondominated.

Now we have a criterion $\varphi$ evaluating the cost of any Lorenz-vector $L(x)$, we want to determine the optimal solution paths according to $\varphi$. Actually, the algorithm introduced in Section 3 can easily be modified so as to determine solution-paths minimizing $\varphi$. It is sufficient to modify our priority rule by setting $f(n) = \arg\min_{x \in F(n)} \varphi(L(x))$ for any open node $n$. Efficiency can be improved by a slight modification of our pruning rule. We have to prune any subpath having a cost-vector $x$ such that $\varphi(L(x)) > \varphi(L(p))$ for some solution-path $p$ already detected.

**Example 4.** *Consider the graph of Figure 1 and assume that the decision criterion $\varphi(.)$ is such that $\varphi(\ell_0) = 0$,*

$\varphi(\ell_1) = 0.9$ and $\varphi(\ell_2) = 1$ (so that the weights $w_i = \varphi(\ell_i) - \varphi(\ell_{i-1})$ add up to 1). It is easy to check that the labels are expanded in the lexicographic order used in the priority rule introduced in Section 3. Therefore, the trace of the algorithm is similar to the one given in Table 2 until iteration 3. At this stage the solution-path 1 is detected with a Lorenz vector $(9,18)$ such that $\varphi(9,18) = 9(2\varphi(\ell_1) - \varphi(\ell_2)) + 18(\varphi(\ell_2) - \varphi(\ell_1)) = 9$. All the other open labels having a value greater than 9, the algorithm stops. This solution can be seen as pessimistic since focused on the worst case. Note that decreasing the strictly positive difference $w_1 - w_2 = 2\varphi(\ell_1) - \varphi(\ell_2)$ would produce a less pessimistic solution path among Lorenz optima.

## 5 CONCLUSION

We have introduced a formal model for robustness allowing to reflect various behavior patterns towards robustness, depending on the choice of parameters $\varphi(\ell_i)$'s. It should be useful to investigate elicitation methods to construct these parameters. In this respect, transposition of elicitation methods used for assessing utility functions deserves investigation. Another important issue might be to investigate the extension of our work when additional information about the likelihood of scenarios is present. Assume for instance that the probability $p_i$ of each scenario is known, the comparison of two cost-vectors might be performed using second order stochastic dominance, which can be defined from Lorenz dominance using probabilistic cumulative distribution functions as suggested in Moyes (1999). Finally, notice that the idea of robust solution and the use of L-dominance is worth studying in other contexts in Artificial Intelligence like planning, valued constraint-satisfaction problems and game search (Dasgupta et al., 1996b), but also in combinatorial problems studied in Operations Research (see e.g. the concept of equitable solutions in Ogryczak (2000)).

**Acknowledgements**

We wish to thank Jean-Yves Jaffray for helpful comments on an earlier version of this text.